\documentclass[preprint,12pt]{elsarticle}

\usepackage{amssymb}

\journal{.}

\newcommand{\ceckd}{CEC{\tiny KD}}

\begin{document}

\begin{frontmatter}



\title{Indexing the Event Calculus with Kd-trees to Monitor Diabetes}


\author[label1]{Stefano Bromuri \corref{cor1}}
\ead{stefano.bromuri@ou.nl}
\cortext[cor1]{Corresponding Author}

\author[label2]{Albert Brugues De la Torre}

\author[label2]{Fabien Duboisson}

\author[label2]{Michael Schumacher}

\address[label1]{Open University of the Netherlands,\\ Management, Science and Technology,\\ Valkenburgweg 177, 6419 AT, Heerlen, Netherlands}

\address[label2]{University of Applied Sciences Western Switzerland\\
	Institute of Business Information Systems\\ TechnoArk 3,\\ CH-3960, Sierre, Switzerland\\ \phantom{a}}

\begin{abstract}

Personal Health Systems (PHS) are mobile solutions tailored to monitoring patients affected by chronic non communicable diseases. A patient affected by a chronic disease can generate large  amounts of events. Type 1 Diabetic patients generate several glucose events per day, ranging from at least 6 events per day (under normal monitoring) to 288 per day when wearing a continuous glucose monitor (CGM) that samples the blood every 5 minutes for several days. This is a large number of events to monitor for medical doctors, in particular when considering that they may have to take decisions concerning adjusting the treatment, which may impact the life of the patients for a long time. 
Given the need to analyse such a large stream of data, doctors need a simple approach towards physiological time series that allows them to promptly transfer their knowledge into queries to identify interesting patterns in the data. Achieving this with current technology is not an easy task, as on one hand it cannot be expected that medical doctors have the technical knowledge to query databases and on the other hand these time series include thousands of events, which requires to re-think the way data is indexed.  
In order to tackle the knowledge representation and efficiency problem, this contribution presents the kd-tree cached event calculus (\ceckd) an event calculus extension for knowledge engineering of temporal rules capable to handle many thousands events produced by a diabetic patient. \ceckd\ is built as a support to a graphical interface to represent monitoring rules for diabetes type 1. In addition, the paper evaluates the \ceckd\ with respect to the cached event calculus (CEC) to show how indexing events using kd-trees improves scalability with respect to the current state of the art.

\end{abstract}


\end{frontmatter}


\section{Introduction}
\label{intro}

Chronic non communicable diseases are becoming a big challenge for the contemporary world, in particular because the people affected by such conditions are growing in numbers. This is a positive fact, as it implies that better cures are available to the public, but it also brings the consequence that healthcare costs rise considerably. In addition, doctors get flooded from information produced by the patients, who more and more often make use of personal health systems (PHS) \cite{DBLP:journals/jaise/BromuriPSKRS16} and wearable devices to monitor their own condition. 
PHS research has mostly focused on interconnection problems, namely on transferring the physiological data of the patient to the hospital infrastructure using interoperability standards. A second generation of PHS is though starting to borrow terminology from business intelligence. In particular, it is becoming of major importance to define tools that allow medical doctors to provide a description of the interesting events happening in the stream of information of the patient. Prescriptive medical reasoning, in the form of temporal rules applied to the stream of physiological values of the patient is becoming an interesting field of research.
There are several issues that need to be solved to be able to define dynamic and personalizable knowledge driven approaches to monitor or query the physiological values of a patient.
First of all, there is the need of a formalism, graphical or textual, that would allow medical doctors to specify patterns of interest coming from their own knowledge. Secondly, there is the problem associated with handling long streams of events happening in relatively shorts time spans, as for examples it happens with continuous glucose monitoring (CGM) that produces 288 events per day.

The Event Calculus (EC) is a formalism for reasoning about events and their effects in a computational logic framework. The original EC, pioneered by Kowalski and Sergot~\cite{ec}, has been extended and adapted to support multiple types of applications, ranging from AI planners~\cite{DBLP:journals/jlp/Shanahan00}, 
web-services~\cite{DBLP:journals/isci/OzorhanKC10}, multi-agent environment platforms~\cite{golemdebs},  and activity recognition systems \cite{Artikis:2010:LPA:1877937.1877941}. 

A widely adopted version of the EC is the version presented in~\cite{DBLP:books/sp/wooldridgeV99/Shanahan99}, which is often interpreted under the semantics of normal logic programs with negation as failure. What is appealing with this version is that the developer specifies which fluents are initiated and terminated by domain specific events, and it is then the domain independent axioms of the EC that cater for which fluents hold at different times. Despite the simplicity of this EC version, the domain independent axioms are computationally naive for large narratives, which can lead to impression that common EC specifications do not scale up.

In~\cite{ChittaroM96}, Chittaro and Montanari study the computational complexity of EC within normal logic programs. Their proposal involves a Cached Event Calculus (CEC) that caches the maximum validity intervals (MVIs) for a fluent by moving some of the computational complexity from query to update time. 
The results achieved by~\cite{ChittaroM96} are important as they also clearly show the theoretical complexity of the CEC in health monitoring settings \cite{DBLP:conf/aime/ChittaroRD95}, showing that CEC is suitable for monitoring applications. One drawback of the current CEC, and of any EC dialect using Prolog, is that it relies on the indexing capabilities of a Prolog engine, which typically uses hash maps to index on the functor name and first argument of the facts stored in a logic based knowledge base. Such an indexing mechanism is inflexible as the application developer cannot change it, as in database systems. 

The indexing problem is worsened by applications, such as patient monitoring, producing large event narratives .
In this paper we propose \ceckd,\ an integration of the CEC with an alternative indexing mechanism for events and MVIs using Kd-trees~\cite{kdtrees}. A Kd-tree is a space-partitioning data structure for organizing points in a k-dimensional space. Queries and updates in the Kd-tree operate on a hyper-plane region containing the multidimensional points and have tractable computational properties~\cite{computationalgeometrybook}. The contribution of the integration of CEC with Kd-trees in \ceckd\ is twofold: (a) we study theoretically the computational aspects of \ceckd\ knowledge representation, showing how this is an improvement with respect to CEC; and (b) we experimentally validate the complexity of \ceckd\ specifications for monitoring applications that require long narratives, putting ourselves in comparison with the state-of-the-art approach represented by CEC.

The research presented in this paper, takes place within the D1NAMO project, a Swiss national project aiming at monitoring patients affected by diabetes type 1 in order to define a non invasive PHS solution to monitor and understand hypoglycaemia events. The algorithms presented in this paper are therefore partially evaluated on D1NAMO data and partially evaluated through a simulation.

The reminder of this paper is structured as follow: Section \ref{sec:rulemag} presents the graphical rule editor of D1NAMO and how this uses EC to calculate alerts concerning the health of the patient; Section \ref{sec:manyevents} presents \ceckd\ and its indexing mechanism; Section \ref{sec:evaluation} evaluates \ceckd\ performance with real events produced in the D1NAMO project; Section \ref{sec:relwork} discusses relevant related work; finally Section \ref{sec:conc} concludes this work discussing about potential future work.

\section{Rule Management in D1NAMO}
\label{sec:rulemag}

Data analysis is becoming more and more an important feature for medical systems and thus this is important also from D1NAMO perspective. Most of data analysis systems include a component called datawarehouse. A datawarehouse is a component that, contrarily to a relational data base, takes the data an put it in a format that is convenient for applying data analysis algorithms. 
In this sense, D1NAMO datawarehouse serves as an event feeds for our rule management approach.

The Rule interface of D1NAMO is a component that interfaces with the datawarehouse of D1NAMO to collect the data of the patient. Effectively speaking we consider a Web service model in which the Rule interface is notified of changes by means of a REST interface receiving the CSV files in input. The CSV files are then saved in local by the rule interface for the data analysis. Thus, the CSV files received from D1NAMO datawarehouse are translated into events which then can be queried using D1NAMO rule interface.

D1NAMO rule interface makes use of a formalism called Event Calculus (EC) \cite{ec}.
EC is based on a many-sorted first-order predicate calculus represented as logic programs that are executable in Prolog. The underlying time model is linear. The EC manipulates fluents.  
A fluent represents a property which can have different values over time.

\begin{table}[htbp!]
	\begin{center}
		{\scriptsize
			\begin{tabular}{|l|p{4cm}|}
				\hline
				{\bf Predicate} & {\bf Meaning}\\ \hline
				initially(F=V)  & The value of fluent {\tt F} is {\tt V} at time 0. \\ \hline
				holds\_at(F=V,T) & The value of fluent {\tt F} is {\tt V} at time {\tt T}. \\ \hline 
				holds\_for(F=V,[Tmin,Tmax]) & The value of fluent {\tt F} is {\tt V} between time {\tt Tmin} and time {\tt Tmax}. \\ \hline 
				initiates\_at(F=V, T ) & At time {\tt T} the fluent {\tt F} is initiated to have value {\tt V}. \\  \hline
				terminates\_at(F=V, T ) & At time {\tt T} the fluent {\tt F} is terminated from having the value {\tt V}. \\  \hline
				broken(F=V, [Tmin, Tmax]) & The value of fluent {\tt F} is either terminated at {\tt Tmax}, or initiated to a different value than {\tt V} between {\tt Tmin} and {\tt Tmax}. \\ \hline
				happens\_at(E,T) & An event {\tt E} takes place at time {\tt T} updating the state of the fluents \\ 
				\hline
			\end{tabular}
		}
	\end{center}
	\caption{EC with multi-valued fluents: predicates.}
	\label{tab:ltarec}
\end{table}

The term $F=V$, denotes that fluent $F$ has value $V$ that has been initiated by an action at
some earlier time-point and not terminated by another action in the
meantime. Tab.~\ref{tab:ltarec} summarizes the main EC predicates we use in this contribution.

Predicates, function symbols and constants start with a lower-case letter while variables (starting with an upper-case letter) are universally quantified.

The specifications of the axioms of the EC are then represented below.

{\scriptsize
	\begin{tabbing}
		(EC0) holds\_at\=(F=V, 0) $\leftarrow$\\ 
		\>initially(F=V).  \\
		\\
		(EC1) holds\_at(F=V, T) $\leftarrow$\\ 
		\>initiates\_at(F=V, T$_s$),\\  
		\>Tmin $<$ T,\\
		\>not broken(F=V, [Tmin, T]).\\
		\\                            
		(EC2) broken(F=V$_1$, [Tmin,Tmax])$\leftarrow$\\  
		\>(terminates\_at(F=V$_1$,T)\\
		\>Tmin $<$ T, Tmax $>$ T);\\
		\>(initiates\_at (F=V$_2$,T), \\
		\>V$_1 \neq$  V$_2$, \\ 
		\>Tmin $<$ T, Tmax $>$ T).\\ 
		(EC3) initiates\_at(F=V, T) $\leftarrow$ \\
		\>happens\_at(Ev,T), \\
		\>Conditions[T].\\
		\\
		(EC4) terminates\_at(F=V, T)$\leftarrow$\\
		\>happens\_at(Ev,T), \\
		\>Conditions[T].\\
		\\
		(EC5) holds\_for(F=V, [Tmin,Tmax])$\leftarrow$\\
		\>initiates\_at(F=V,Tmin)\\ 
		\>terminates\_at(F=V,Tmax)\\ 
		\>End $>$ Start, not broken(F=V, [Tmin,Tmax]).\\ 
		\\
		(EC6) holds\_for(F=V, [Tmin,infPlus])$\leftarrow$\\
		\>initiates\_at(F=V,Tmin),\\ 
		\>not broken(F=V,[Tmin,infPlus]).\\ 
		\\
		(EC7) holds\_for(F=V, [infMin,Tmax])$\leftarrow$\\
		\>terminates\_at (F=V,Tmax),\\ 
		\>not broken(F=V,[infMin,Tmax] ).\\

	\end{tabbing}  
}

Clause EC0 states that a property {\tt F} holds at time {\tt 0} if
an {\tt intially/1} predicate is true at time {\tt 0}.
Clause EC1 states that a property holds at a time {\tt T} if it has been initiated at time {\tt Tmin} and the holding of that property has not been broken between the starting time {\tt Tmin} and
the time of interest {\tt T}. To decide when a property is broken, we use
the clause EC2. This states that a property {\tt P} is broken between time
{\tt Tmin} and {\tt Tmax}, if it is terminated at a time {\tt T} between {\tt Tmin} and {\tt Tmax} or initiated to a different value between {\tt Tmin} and {\tt Tmax}.  
\begin{figure}[h]
	\centering
	\includegraphics[width=.4\textwidth]{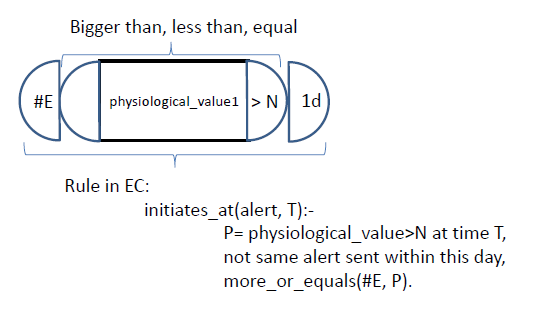}
	\includegraphics[width=.4\textwidth]{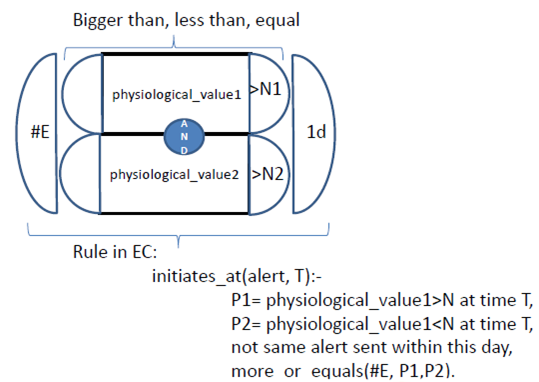}
	\caption{Simple Pattern Specification and Complex Pattern Specification.}
	\label{fig:simple}
\end{figure}
The other clauses specify when a property is initiated
(EC3) or terminated (EC4), in terms of the conditions holding in the
current context, typically expressed in terms of the {\tt holds\_at/2}, {\tt holds\_for/2}
predicates, meaning that such clauses will change according to the particular
domain being modeled with the EC. EC5-EC7 express the EC clauses that deals with validity intervals of fluents. In particular, EC5 specifies that a fluent {\tt F} keeps a value {\tt V} for an interval going from {\tt Tmin} to {\tt Tmax} if nothing happens in the middle that breaks such an interval. EC6-EC7 behave like EC5, but deal with open intervals.

\begin{figure}[h]
	\centering
	\includegraphics[width=.7\textwidth]{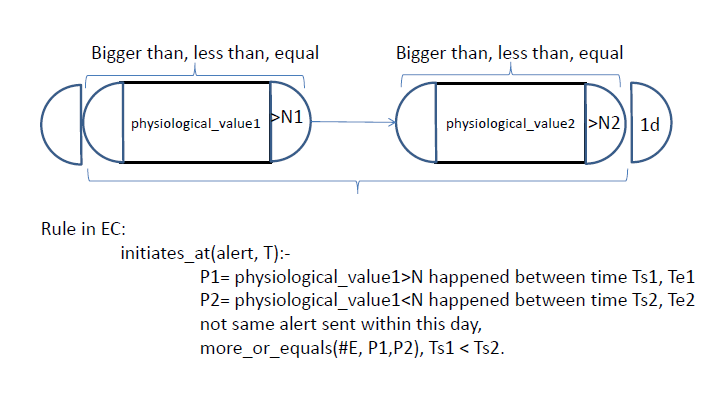}
	\includegraphics[width=.5\textwidth]{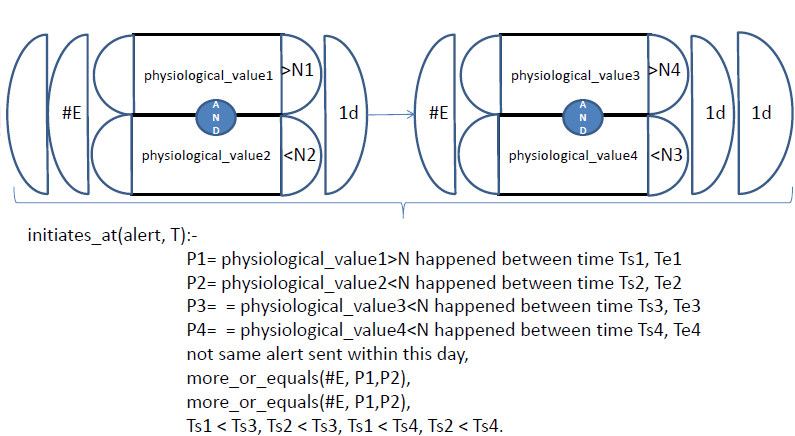}
	\caption{Sequential Pattern Specification and Complex Sequential Pattern Specification.}
	\label{fig:sp}
\end{figure}

The rule interface builds on top of the EC to specify the following types of patterns, ordered by complexity:
\begin{enumerate}
	\item Simple Pattern: as shown in Fig. \ref{fig:simple} a simple pattern simply consider a threshold value, like for example on glucose values to be violated a certain number of times within a time period.
	\item Complex Pattern: as shown in Fig. \ref{fig:simple} a complex pattern considers the occurrence of several threshold
	violations concerning multiple signals within a certain time period.
	\item Sequential Pattern: as shown in Fig. \ref{fig:sp} sequential patterns show the occurrence of a threshold violation
	followed by another threshold violation.
	\item Complex Sequential Pattern: as shown in Fig. \ref{fig:sp} complex sequential patterns are a combination of sequential patterns and complex patterns.
\end{enumerate}

The predicates shown in Figs \ref{fig:simple}-\ref{fig:sp} depend on a set of meta-predicates to evaluate the execution
of the logical patterns and put it in relationship with the other patterns. In particular, we modelled {\textsc more\_or\_equals/3}:

{\scriptsize
	\begin{tabbing}
		more\=\_or\_equals\_to(Frequency, Pattern, Time)$\leftarrow$\\
		\> apply(Pattern,Time)\\
		\> findall(\_, Pattern,List), size(List,S), S=Frequency.
	\end{tabbing}}
	
	the pseudocode above takes the temporal pattern {\sf Pattern} and it unifies it with the time variable {\sf Time}, after this the {\sf findall/3} predicate executes the pattern to see how many times it holds in the specified time.
	A complex pattern then uses the {\textsc more\_or\_equals/3} predicate on multiple patterns, using the declarativity of Prolog to combine the patterns. Sequential patterns use a {\textsc constrained\_more\_or\_equals/5} predicate, in particular this predicate takes into consideration the execution of the patterns between two periods of time that must be strictly happening in a sequential order.

	{\scriptsize
		\begin{tabbing}
			constrained\_more\=\_or\_equals\_to(Frequency, Pattern, Pattern2, Period1, Period2)$\leftarrow$\\
			\> apply(Pattern,Period1),\\
			\> apply(Pattern2,Period2),\\
			\> findall(\_, (Pattern,Pattern2,List), size(List,S),\\ 
			\>S=Frequency.
		\end{tabbing}}

\begin{figure}[h]
			\centering
			\includegraphics[width=1\textwidth]{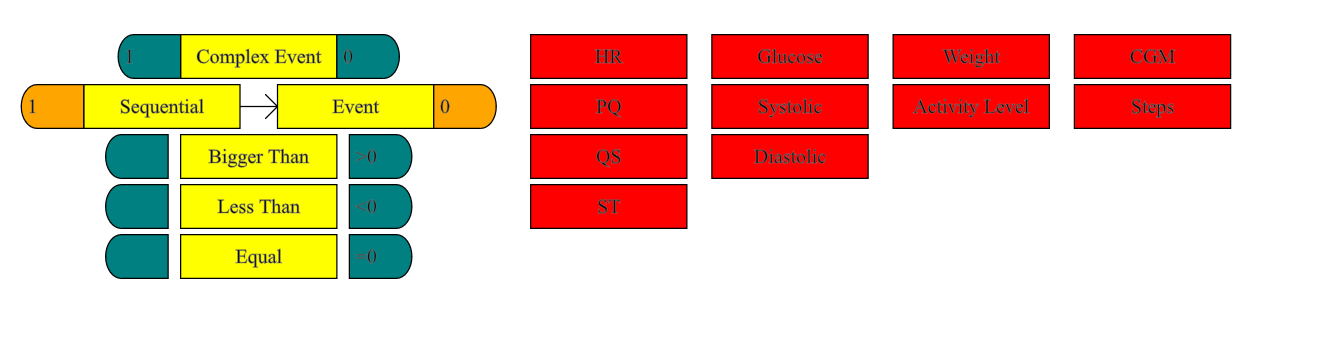}
			\caption{Javascript Rule Interface in D1NAMO.}
			\label{fig:din}
		\end{figure}

		The definitions of {\textsc more\_or\_equals/3} and {\textsc constrained\_more\_or\_equals/5} are not recursive because we found that the semantics become unclear to a reader which is something we explicitly want to avoid given the medical settings of D1NAMO. Thus, to maintain understandability of the generated rules, we decided to allow only for a double level of nesting of the rules. Fig. \ref{fig:din} provides an example of a JavaScript implementation of the rule interface of D1NAMO.

\begin{figure}[htb!]
			\centering
			\includegraphics[width=.7\textwidth]{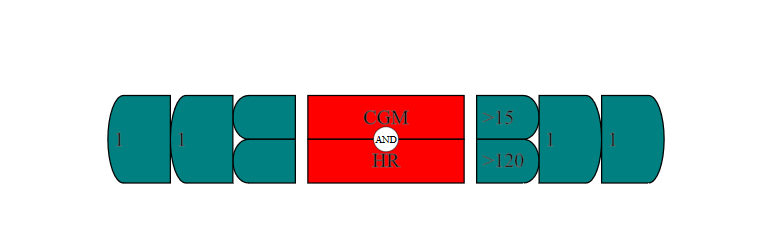}
			\caption{Rule 0 Specification in the Graphical Rule Editor.}
			\label{fig:lul}
		\end{figure}

		The rules created with the rule editor are compiled directly into {\sf initiates\_at/2} predicates.
		For example, if we wanted to specify a pattern to monitor continuous glucose events and heart rate at
		the same time, we could use a complex pattern as shown in Fig. \ref{fig:lul}. 
		
		The literal meaning of the {\sf rule0} pattern shown in Fig. \ref{fig:lul} is to report on each complex pattern in which, within
		a time window of one day, CGM measurements are above 13 and Heart rate is above 130. The rule editor of D1NAMO
		translates this pattern into the rule specified below, by using ECG predicates.
		
		{\scriptsize
			\begin{tabbing}
				initiates\=\_at(generic\_alert([doctor, rule0])=up(normal,rule0), T):-\\
				\>Tbefore is T-1,\\
				\>holds\_at(obs(cgm)=value(ValCGM),T),\\ 
				\>ValCGM$>$13.0,\\
				\>holds\_at(obs(hr)=value(ValHR),T), \\
				\>ValHR$>$120.0,not((query\_kd(happens\_at(sent\_alert(generic\_alert([doctor]), Th), [Tbefore,T])))). \\
				\\
				initiates\_at(generic\_alert([doctor,rule0])=sent, T):-\\ 
				\>happens\_at(sent\_alert(generic\_alert([doctor,rule0])),T).
			\end{tabbing}
		}
		
		Notably, the patterns are all transformed in alerts that are then reported in the editor after deployment and execution of the rule.
		After the rule is deployed, the editor can be used to play events and effectively highlight patterns of interest happening the stream
		of data of the patient.

		\begin{figure}[htb!]
			\centering
			\includegraphics[width=.7\textwidth]{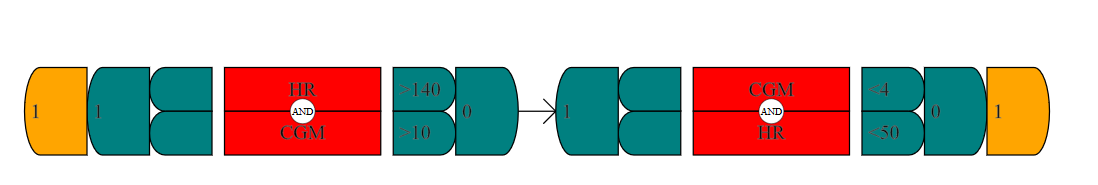}
			\caption{Rule 1 Specification in the Graphical Rule Editor.}
			\label{fig:lul2}
		\end{figure}

		The pattern specified in {\sf rule1} in Fig. \ref{fig:lul2} has a more complex interpretation. 
		Such a pattern translates to the following EC theory:
		
		{\scriptsize
			\begin{tabbing}
				initiates\=\_at(generic\_alert([doctor])=up(normal,rule1), T):-\\
				\>Tbefore is T-1, \\
				\>more\_or\_equals\_to(1, (hr$>$130.0,cgm$>$15.0),[Tbefore, T],TimeConstraints),\\ 
				\>constrained\_more\_or\_equals\_to(1, (cgm$<$5.0,hr$<$60.0),[Tbefore,T],TimeConstraints),\\
				\>not((query\_kd(happens\_at(sent\_alert(generic\_alert([doctor]), Th), [Tbefore,T]))). \\
				\\	
				\>initiates\_at(generic\_alert([doctor,rule1])=sent, T):- \\
				\>happens\_at(sent\_alert(generic\_alert([doctor,rule1])),T).
			\end{tabbing}}
			
			The pattern states to look for complex pattern in which a heart rate of 130 and a CGM value above 15 is then followed by a drop in CGM and 
			heart rate within time frame of one day. Such a pattern can be used in D1NAMO to observe patience that present a variable diabetes pattern.
			The main issue with defining these patterns is that they depend on primitives of the EC such as the {\sf holds\_at/2} predicate. The main issue of EC predicates is that, being based on declarative programming and the {\sf negation as failure} procedure of Prolog, they may be quite expensive to compute.
			For this reason, the next session presents the \ceckd event calculus, an extension of the cached event calculus \cite{cecmedicine} (CEC) that indexes the events using kd-trees structures.

			\section{Dealing with Many Events in the Event Calculus}
			\label{sec:manyevents}
			With respect to the rules that we model for D1NAMO, the EC formalism gets particularly slow
			when dealing with large amounts of events. Recent research has tried to overcome this issue in several ways \cite{DBLP:journals/tkde/ArtikisSP15,cecmedicine}, but the amount of events that can be processed with such approaches is still very limited and not adequate to the issue of monitoring patients affected by a chronic illness, who notably produce many thousands of events even within one week. If we consider continuous glucose monitoring, for example, at least 288 values per day are produced. The normal versions of the EC currently available in research would not be able to cope with such a large number of events without becoming a bottleneck for the analysis to be performed. Within this deliverable we present and evaluate a solution for this problem. We described our extended EC as the \ceckd\ event calculus, as it caches events like the cached event calculus, but it also indexes them using KD-trees. 
			Such an event calculus has been defined with the idea of dealing with a large number of events. The necessity to define this extension resides in the fact that the medical domain typically implies heterogeneous types of events. For example, in D1NAMO we are faced with discrete events (point of care glucose samples) and continuous events (ECG, activity, continuous glucose monitors). As a consequence, to have a tool that can be truly useful, the temporal analysis must scale to thousands of events of different kind. The knowledge representation of the EC framework allows us an easy inclusion of predicates to query different physiological values. The creation of a quick indexing scheme, as described in this deliverable, allows to deal with discrete and continuous domains.

			\subsection{K-dimensional Trees}
			
\begin{figure}[htp]
				\centering
				\includegraphics[scale=0.205]{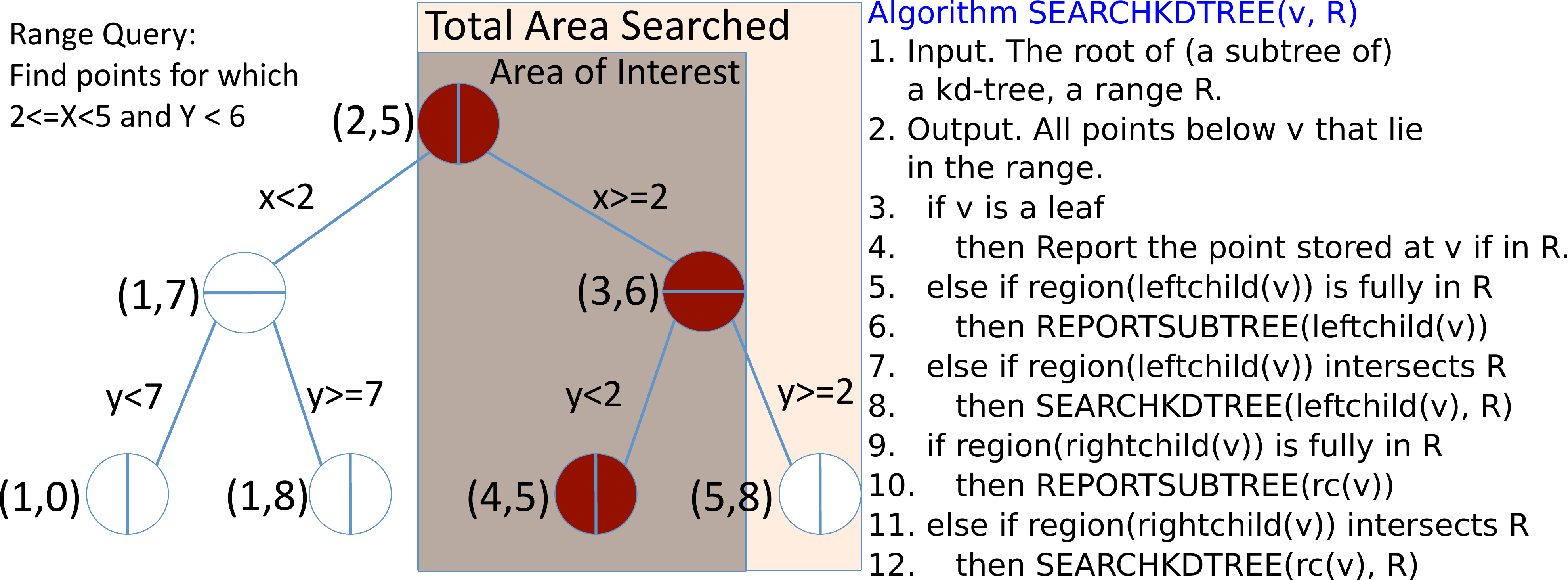}
				\vspace{-0.2cm}
				\caption{Range Query on a Kd-tree~\cite{computationalgeometrybook}}
				\vspace{-0.6cm}
				\label{query1}
			\end{figure}

			Kd-trees \cite{kdtrees} are binary trees optimised to deal with k-dimensional points.
			As reported in \cite{computationalgeometrybook}, given a set of k-dimensional points, we 
			can generate a Kd-tree by splitting recursively the hyperplane containing the points at every level of the tree, alternating the 
			coordinate that is split according to the depth of the tree. Fig.~\ref{query1} shows how splits are performed on
			a 2-dimensional tree of depth 3, where at each level the value of the splitting coordinate is the median value, deciding if a new point should go to the left or to the right of an existing tree node.

			\noindent  Fig. \ref{query1} shows also the effect of searching a Kd-tree via a range query performed on it. The range query algorithm recursively searches for regions contained or intersected by the region specified in the range query. If the region found is contained in the region specified in the query, then the whole region is returned. If the region of the tree intersects the region specified by the query, then the points reported are only those ones included in the region of the query.   
			
			The Kd-tree data structure has a set of important properties when dealing with searches of multi-dimensional points:
			(a) a Kd-tree for a set P of n points uses $O(n)$ storage 
			and can be constructed in $O(nlog(n))$ time; 
			(b) the operations of adding or deleting a point have a complexity of
			$O(log(n))$; (c) a rectangular range query on the Kd-tree takes
			O($\sqrt{n} + k$) time, where $k$ is the number of reported points residing the rectangular area identified by the query.
			
			These properties are fundamental to create a version of the EC that can scale up to be used in dynamic applications
			with large narratives. 
			
			\subsection{The Cached Event Calculus with Kd-Trees}
			\label{sec:oeckdcomb}
			
			\begin{table}[htb!]
				\begin{center}
					{\scriptsize
						\begin{tabular}{|p{4.8cm}|p{5cm}|p{3.2cm}|}
							\hline
							{\bf Primitive Operations to Manage the Tree} & {\bf Meaning} & {\bf Theoretical Complexity}\\ \hline
							create\_kdi(+L, -Idx)/destroy\_kdi(+L) & Creates/Destroys a four-dimensional Kd-tree index {\sf Idx} identified by label {\sf L} & $O(1)$. \\ \hline
							kdi(+L, -Idx) & Returns an existing four-dimensional Kd-tree index {\sf Idx} identified by label {\sf L}& $O(1)$. \\ \hline
							insert\_kdi(+Idx, (+Arg$_1$,+Arg$_2$,+Arg$_3$,+Arg$_4$),Value)/ delete\_kdi(+Idx, (+Arg$_1$,+Arg$_2$,+Arg$_3$,+Arg$_4$))  & Inserts a four-dimensional key, whose coordinates are {\sf Arg$_1$}, {\sf Arg$_2$}, {\sf Arg$_3$}, {\sf Arg$_4$}, associated to a value {\sf Value} in a Kd-tree index {\sf Idx}. Such a predicate
							transforms the arguments in long integer values using an hashing function, whose range can be between $-\infty$ and $+\infty$. delete\_kdi/2 deletes a value given a four dimensional key like for insert\_kdi/2.  & $O(log(|Events|))$ for the events tree,  $O(log(|MVI|))$ for the MVI tree. \\ \hline
							range\_query\_kdi(+Idx, (?R$_1$,?R$_2$,?R$_3$,?R$_4$), -Result) & Produces a four dimensional rectangular range query on the four dimensional kd-tree {\sf Idx}, where the query is specified by the ranges {\sf R$_1$}, {\sf R$_2$}, {\sf R$_3$}, {\sf R$_4$} and the result is unified with the variable {\sf Result}. The range arguments can be fixed values or specified as [StartValue, EndValue]. & $O(\sqrt(|MVI|) + K_{F,V})$ where MVI is the total number of MVIs  and $K$ is a constant related to the number of points reported depending on the value of  F and  V. \\ \hline
							\hline
							{\bf Cache Operations for Events and MVIs Indexing} & {\bf Meaning} & {\bf Theoretical Complexity} \\ \hline
							update(+Ev,+T) & Indexes an event happening at time {\sf T} in a four dimensional Kd-tree index and caches its consequences in a MVI Kd-tree index. & $O(n(\sqrt(|MVI|) + K_{F,V}))$. Where $n$ is the number of queries to the context in the initiates\_at/3/terminates\_at/3 EC rules.\\ \hline
							index(+Ev,+T) & Indexes an event happening at time {\sf T} in a Kd-tree index.  & like insert\_kdi/5. \\ \hline
							cache(+Ev,+T) & Caches an event happening at time {\sf T} in a MVI Kd-tree index. & like insert\_kdi/5. \\ \hline
							close\_interval(+Idx, (+F,+V,T$_{end}$))/ \hphantom{aaaaaaaaaaaaaaaa}
							open\_interval(+Idx,(+F,+V,T$_{start}$)) & close\_interval/2 Indexes a closed MVI whose fluent is {\sf F}, whose value is {\sf V}, and whose ending time is {\sf T$_{end}$} in {\sf Idx}.  open\_interval/2 is like {\sf close\_interval/2}, but the time indexed is the starting time {\sf T$_{start}$}&  
							$O(log(|MVI|))$. \\ \hline
							intersect\_query(+Idx, (?F,?V,-T$_{s}$,-T$_{e}$,+WT$_s$,+WT$_{e}$)) & Uses range queries to find the MVIs that intersect WT$_s$,WT$_{e}$ and unifies with {\sf F},{\sf V},{\sf T$_s$}, {\sf T$_e$}. &  $O(\sqrt(|MVI|) + K_{F,V})$.\\ \hline
							cached\_between(+WT$_s$,+WT$_{end}$, mholds\_for(?F=?V,[?T$_s$,?T$_{end}$])) & Queries for the MVIs intersecting the time window between WT$_s$ and WT$_{e}$. & $O(2*\sqrt(MVI) + K_{F,V})$.\\ \hline
							
						\end{tabular}}
					\end{center}
					\caption{Predicates of \ceckd. The symbols +, - and ? indicate respectively inputs, outputs and inputs/outputs.}
					\label{tab:CECkd}
				\end{table}
				
				To obtain a version of the CEC that can scale up with respect to the number of events, we use two four-dimensional Kd-trees, one to index the events and one to index the MVIs in which a fluent holds. 
				In addition to the CEC predicates, in \ceckd~we introduce new predicates to model
				our knowledge as stored in the Kd-trees. Table \ref{tab:CECkd} summarises the predicates of \ceckd~and their theoretical complexities.
				These predicates take a query produced on an event or an MVI and translate the query to an insert point, delete point or range query on the 
				event tree or on the MVI tree. In particular we do so maintaining a declarative approach, to keep the expressivity of CEC intact.
				We start with how we specify the addition of an event using the {\sf update/2} predicate, which is specified as follows:
				
				{\scriptsize
					{\sf
						\begin{tabbing}
							up\=date(Ev,T)$\leftarrow$ index(Ev, T), cache(E, T).\\ 
							\\
							index(Ev,T)$\leftarrow$functor(Ev,Name,Arity), argument(Ev,1, Argument),\\ 
							\>kdi(happens\_at, EvIndx), insert\_kdi(EvIndx,(Name,Arity,Argument,T),Ev).\\
							\\
							cache(Ev,T)$\leftarrow$kdi(mholds\_for, MVIIdx)\\ 
							\>foreach(terminates\_at(Ev,F$_1$=V$_1$,T),close\_interval(MVIIdx,(F$_1$,V$_1$,T))),\\
							\>foreach(initiates\_at(Ev,F$_2$=V$_2$,T),open\_interval(MVIIdx,(F$_2$,V$_2$,T))).
						\end{tabbing}
					}}
					
					To update the knowledge base with an event, we first index the event in a Kd-tree and then we add what is initiated and terminated.
					In \ceckd~the {\sf index/2} predicate in {\sf update/2} stores the produced event in a four dimensional Kd-tree, indexing its name, arity, first argument and time when it happened, using the {\sf insert\_kdi/3} predicate. 
					Once the event is stored in the event Kd-tree, \ceckd~considers this as happened. The 
					event is then used in the {\sf cache/2} predicate to query the fluents whose values are initiated and terminated due to the event happening, which are then cached in the {\sf MVIIdx}. 
					MVIs can be open to infinite or closed. We specify {\sf close\_interval/2} as follows:
					
					{\sf 
						{\scriptsize
							\begin{tabbing}
								clo\=se\_interval(MVIIdx,(F,V,T))$\leftarrow$\\
								\>range\_query\_kdi(\=MVIidx, (F,V,[0,+$\infty$],+$\infty$), mholds\_for(F,V,T$_s$,+$\infty$)), \\
								\>delete\_kdi(MVIidx,(F,V,T$_s$,+$\infty$)),\\
								\>insert\_kdi(\=MVIidx,(F,V,T$_s$,T),mholds\_for(F=V,[T$_s$,T])).
							\end{tabbing}
						}
					}
					\noindent The definition of {\sf close\_interval/2} takes into account the existence of an open interval, that has to be closed due to an event
					terminating the fluent {\sf F} to have value {\sf V}. 
					The procedure to index open intervals is similar to the one for closed intervals with the difference that no MVI is retracted from the {\sf mholds\_for} Kd-tree.
					We then define the {\sf holds\_at/2} predicate, to query the value {\sf V} of a fluent {\sf F} as follows:
					
					{\sf
						{\scriptsize
							\begin{tabbing}
								hol\=ds\_at(F=V,T)$\leftarrow$kdi(mholds\_for,MVIidx),\\
								\> range\_query\_kdi(MVIidx,  (F,V,[0,T],[T,+$\infty$]), mholds\_for(F=V,[T$_s$,T$_e$])).
							\end{tabbing}
						}}
						
						This definition of {\sf holds\_at/2} looks for the interval of time
						that intersects {\sf T}, speeding up the computation of {\sf holds\_at/2}. The {\sf intersect\_query/2} is implemented as a range query 
						on a Kd-tree containing the MVIs, varying the starting time of the MVI from 0 to {\sf T}, and the ending time from {\sf T} to {\sf positive infinite}.
						In the case that also {\sf F} and {\sf V} are variables, the range query is performed also on these coordinates, assuming a range for them between $-\infty$ and $+\infty$. 
						If we want to query for the MVIs of a fluent {\sf F} with value {\sf V} we can redefine the {\sf mholds\_for/2} predicate as 
						follows:
						
						{\sf 
							{\scriptsize
								\begin{tabbing}
									mho\=lds\_for(F=V, [T$_{s}$,T$_{end}$])$\leftarrow$kdi(mholds\_for,MVIidx),\\
									\>range\_query(MVIidx,(F,V,T$_{s}$,T$_{end}$), mholds\_for(F=V,[T$_s$,T$_{end}$])).
								\end{tabbing}
							}
						}
						
						In the worst case, when both {\sf F} and {\sf V} are not defined, the predicate will backtrack through all the MVIs contained in the 
						MVI tree, like the normal CEC would do. In the case that one between F and V is defined, the query will only select those intervals related to a
						particular fluent and return them.
						Furthermore, sometimes it is not worth to query the whole list of MVIs of a fluent, when a fluent is known to
						change quite often in time. To avoid performing expensive queries at update time, we define a further query on the MVI tree as follows:
						
						{\sf 
							{\scriptsize
								\begin{tabbing}
									ca\=ched\_between(WT$_s$,WT$_{e}$, mholds\_for(F=V, [T$_{s}$,T$_{e}$]))$\leftarrow$\\
									\>kdi(mholds\_for,MVIidx),intersect\_query(MVIidx,(F,V,T$_{s}$,T$_{e}$,WT$_s$,WT$_{e}$)).\\
									\\
									in\=tersect\_query(MVIidx,(F,V,T$_{s}$,T$_{e}$,WT$_s$,WT$_{e}$))$\leftarrow$\\
									\>range\_query(MVIidx,(F,V,[0,WT$_{s}$],[WT$_{s}$,$+\infty$]),mholds\_for(F=V,[T$_{s}$,T$_{e}$])).\\
									in\=tersect\_query(MVIidx,(F,V,T$_{s}$,T$_{e}$,WT$_s$,WT$_{e}$))$\leftarrow$\\
									\>range\_query(MVIidx,(F,V,[WT$_{s}$,WT$_{e}$],[WT$_{e}$,$+\infty$]),mholds\_for(F=V,[T$_{s}$,T$_{e}$]))).
								\end{tabbing}
							}
						}
						
						The {\sf cached\_between/3} above uses a time window defined between {\sf WT$_s$} and {\sf WT$_{e}$} to query the MVI tree 
						about the intervals in which a fluent {\sf F} took value {\sf V}. Such a query will return only those intervals which intersect the time window, 
						leaving out the intervals happening before or after the time window. 
						
						With respect to CEC, the main improvement from the perspectivee of computational complexity can be seen in the {\sf update/2} predicate shown in Table \ref{tab:CECkd}. If for CEC, in a context dependent theory, the update time complexity is exponential with respect of the number of context queries performed \cite{ChittaroM96}, for \ceckd~the complexity depends on the number of queries to the context {\sf n} multiplied by the square root of the total number of MVIs and number of MVIs reported by the query. This happens thanks to the fact that we do not rely on negation as failure that would check the whole temporal database for a solution, but we define time windows, which restrict the number of solutions returned, and we use an indexing based on kd-trees, that treat the time windows as range queries. Similar considerations apply for the query time.

						\section{Evaluation}
						
						\label{sec:evaluation}
						
						We evaluated \ceckd~by comparing it with the update time and query time of CEC. In order to perform this comparison we adapted our \ceckd~theory to CEC ensuring the final behaviour of the two theories is the same.
						\begin{figure}[htp]
							\centering
							\includegraphics[width=6cm, height=4cm]{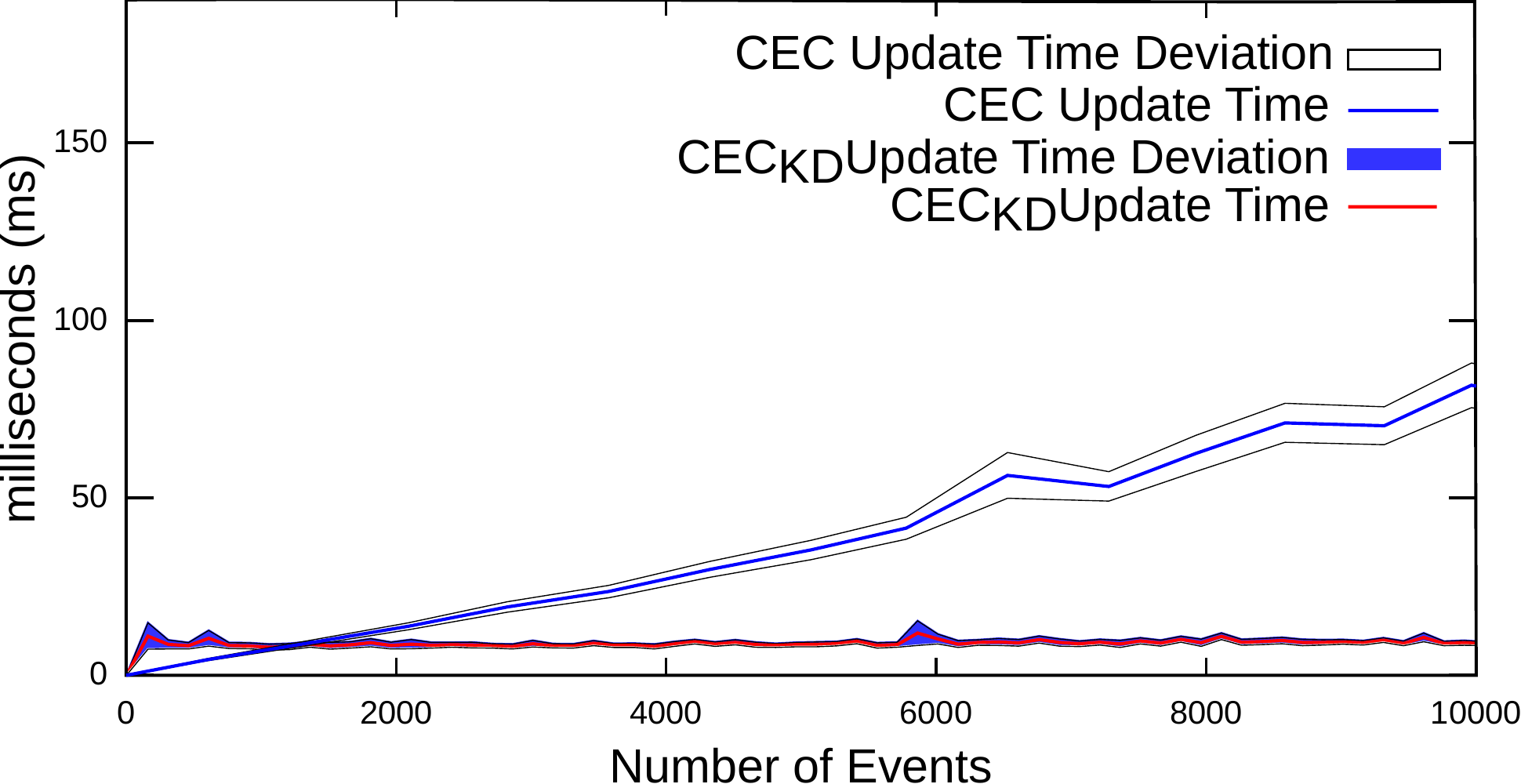}
							\includegraphics[width=6cm, height=4cm]{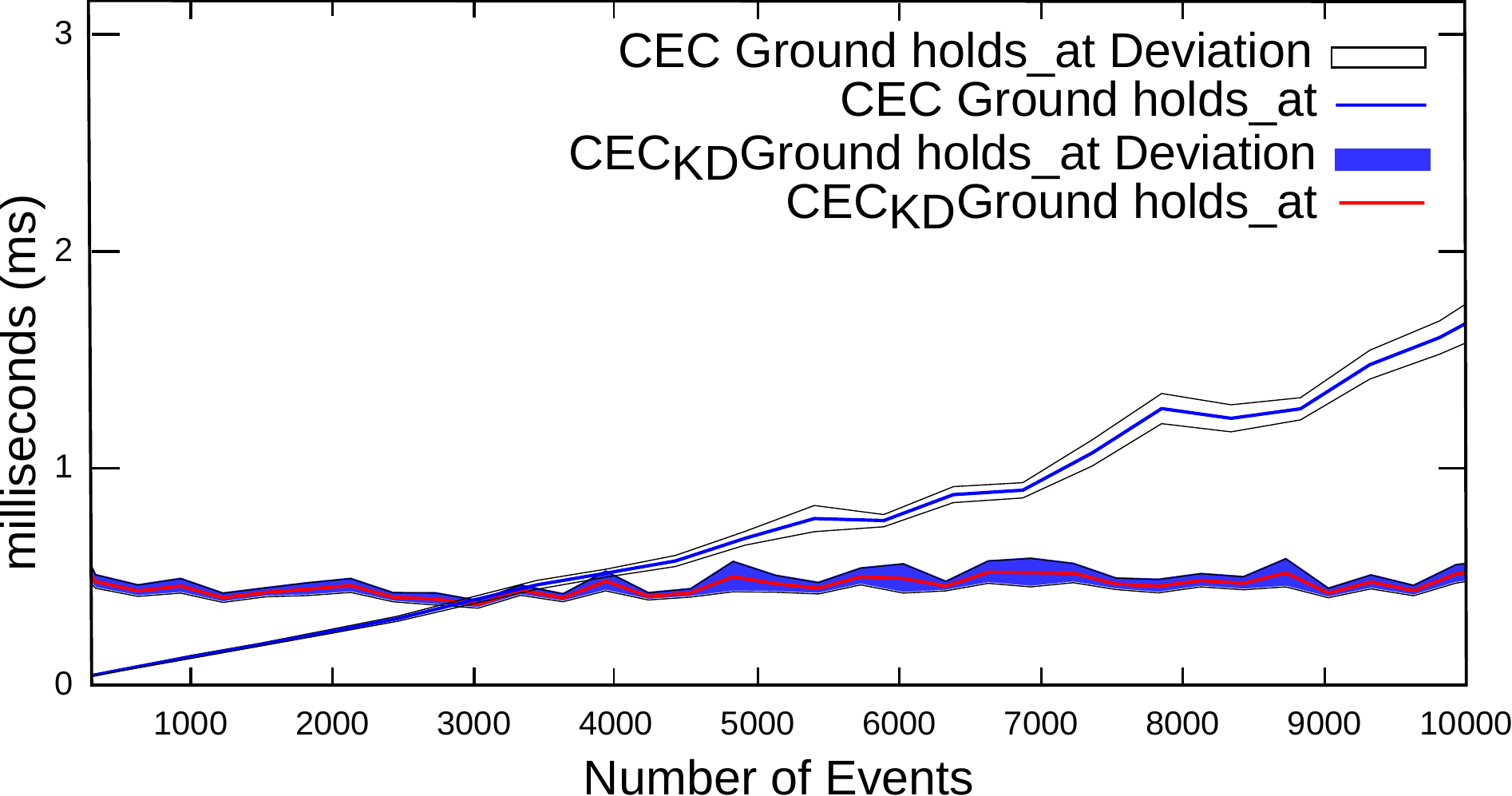}
							\caption{From Top to Bottom: Update Time of CEC vs \ceckd, Results of a Ground holds\_at/2 Query.}
							\label{queryupdate}
						\end{figure}
						Firstly, the testing environment is a Intel Core i7, with 8GBs of RAM and the tests that follows were performed by repeating them 50 times, averaging the results and 
						producing their standard deviation to show their precision. CEC and \ceckd~were executed on 2Prolog version 3.0, using the Java interface of 2Prolog as the basis to connect the Kd-tree structure implemented in Java with Prolog predicates.
						
						For the purpose of the evaluation, we used data collected in the D1NAMO project, this data included 7 patients affected by diabetes type 1, monitored for a total of 355 hours, with a total number of 6660 events.
						
						The number of events that a DT1 patients may produce is variable, they general produce a large number of events within several weeks of observation. In D1NAMO, we did not have event histories longer than 1000 glucose events per patient, so we  bootstraped the available data in order to have event histories of about 10000 events including CGM, normal glucose readings and other events such as weight and recording of meal consumption during the day and created 50 artificial patients. This is acceptable in this settings as we want to evaluate the scalability of the system and not the accuracy of the rules, that in any case are assumed to be dynamically defined by the medical doctors monitoring the patients. 
						
						We then simulated the production of such events and fed them to \ceckd~and CEC. 
						\begin{figure}[htp]
							\centering
							\includegraphics[width=6cm, height=4cm]{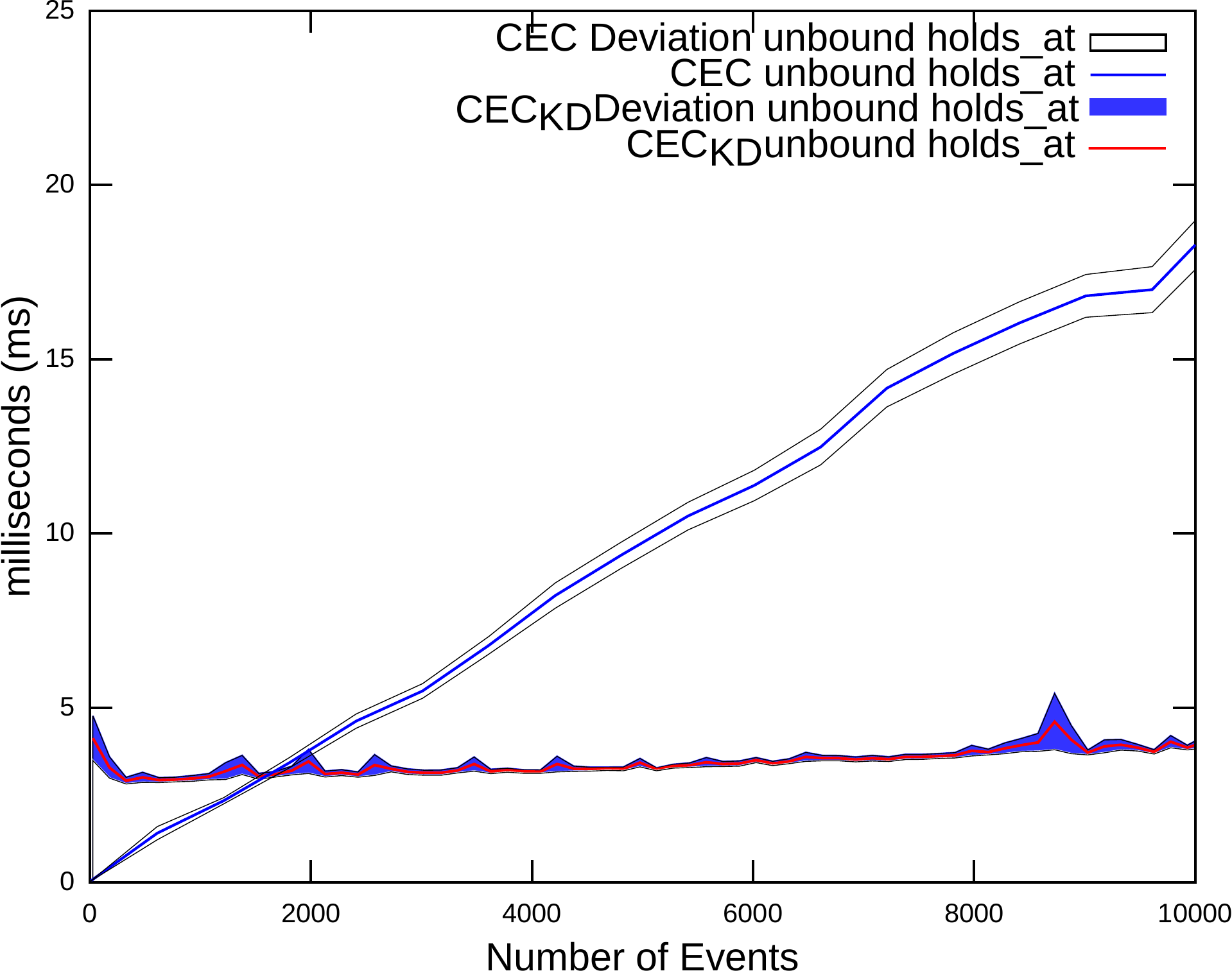}
							\includegraphics[width=6cm, height=4cm]{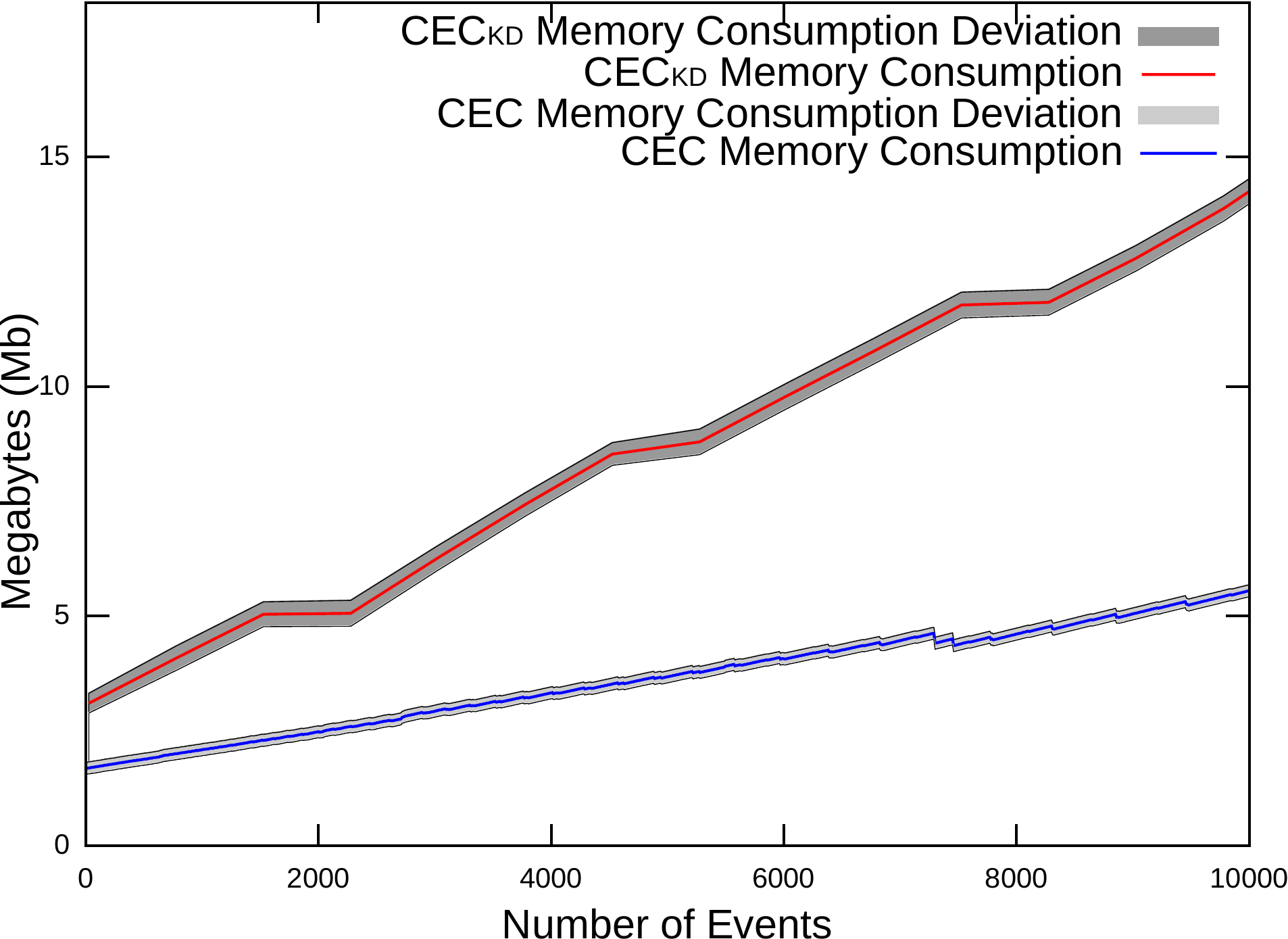}
							\caption{From Top to Bottom: holds\_at/2 Query Completely Unbound and Ram Memory Consumption of CEC vs \ceckd.}
							\label{queryupdate2}
						\end{figure} 
						
						In what follows we compare the update time of \ceckd~and CEC in the DT1 scenario and the {\sf holds\_at/2} predicate for querying the current state of the patient. The update time 
						contains calls to all the predicates of \ceckd~and since {\sf holds\_at/2} predicate is based on {\sf mholds\_for/2} in CEC and  on {\sf range\_query\_kdi/3} in \ceckd, we have a complete picture of both the formalisms.
						The part on top of Fig. \ref{queryupdate} shows the comparison between CEC and \ceckd~with respect to the update time while producing events in the Prolog database.
						On one hand, CEC demonstrates a linear dependency on the number of events produced on the EC database. In particular 2Prolog implements an indexing mechanism on the first term of the dynamic predicates asserted, to speed up the computation, but this is not enough to 
						avoid a linear dependency on the events produced. On the other hand \ceckd~curve has an access time to the Kd-tree structure which does not make it start from 0 nanoseconds, but then it is almost flat throughout the whole simulation. This happens because we make large use of the Kd-tree to avoid computing the whole list of intervals, and by using as much as possible range queries. This results in a computation that depends, for range queries, on the square-root of the events or MVIs stored in the Kd-trees and, for deletion and insertion of events and MVIs, on the logarithm of the number of events or MVIs currently stored. Furthermore the curve associated to \ceckd~looks almost flat because the cost of performing a range query on the Kd-tree is negligible if compared to the cost of performing a linear search on the MVIs stored in 2Prolog, that implies multiple expensive unifications.
						
						The part on the bottom of Fig. \ref{queryupdate} shows the curves resulting from a {\sf holds\_at/2} query where all the arguments are ground. The fact that all the arguments are ground in the query, improves the response of 2Prolog indexing but there is still a linear dependency on the events produced in the database which does not take place with \ceckd. The ground {\sf holds\_at/2} query of \ceckd~is particularly fast thanks to the fact that we can perform a range query where we look for one interval intersecting the current time where the value and the fluent match the query.
						
						\begin{figure}[htp]
							\centering
							\includegraphics[width=6cm, height=4cm]{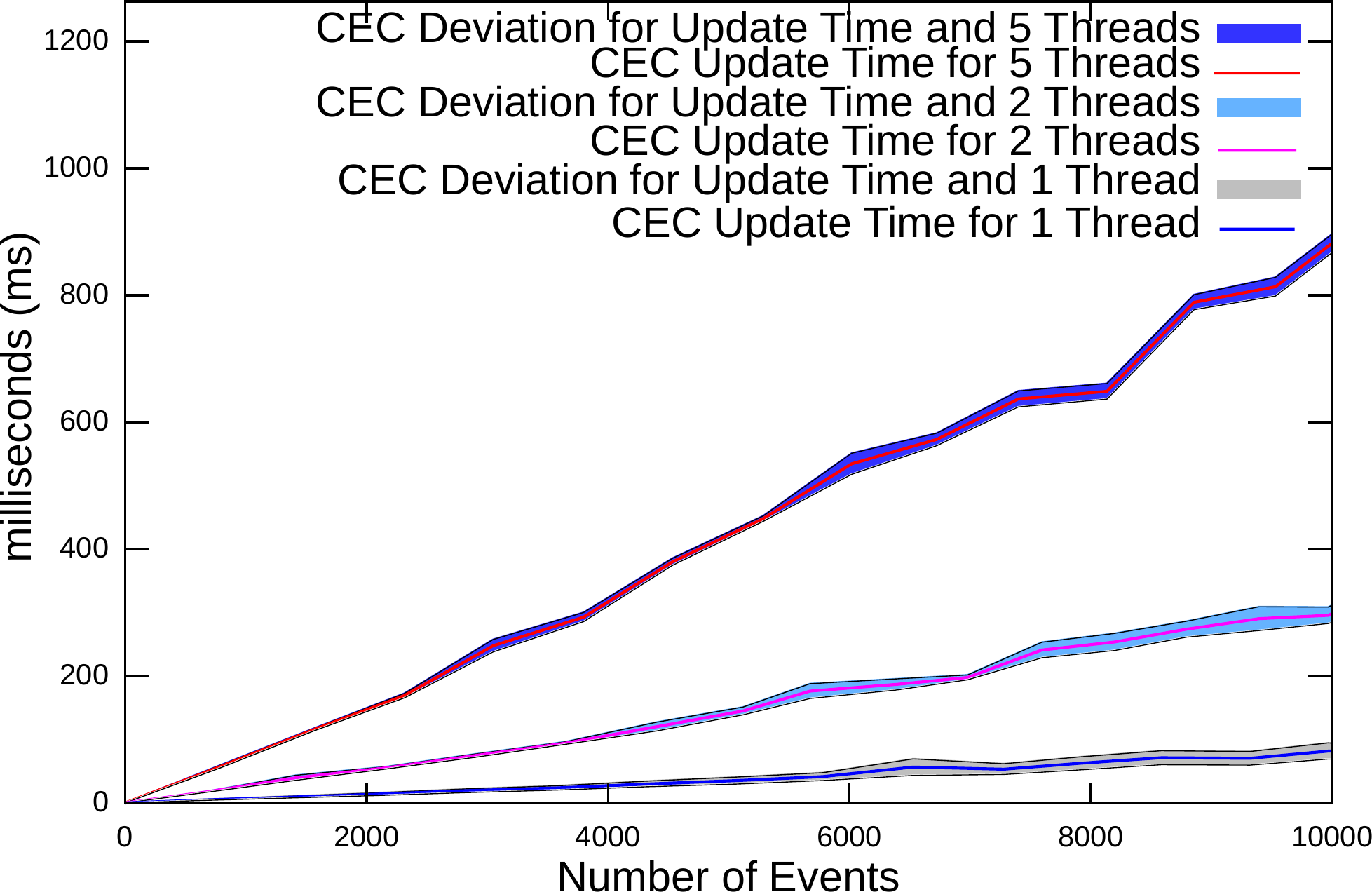}
							\includegraphics[width=6cm, height=4cm]{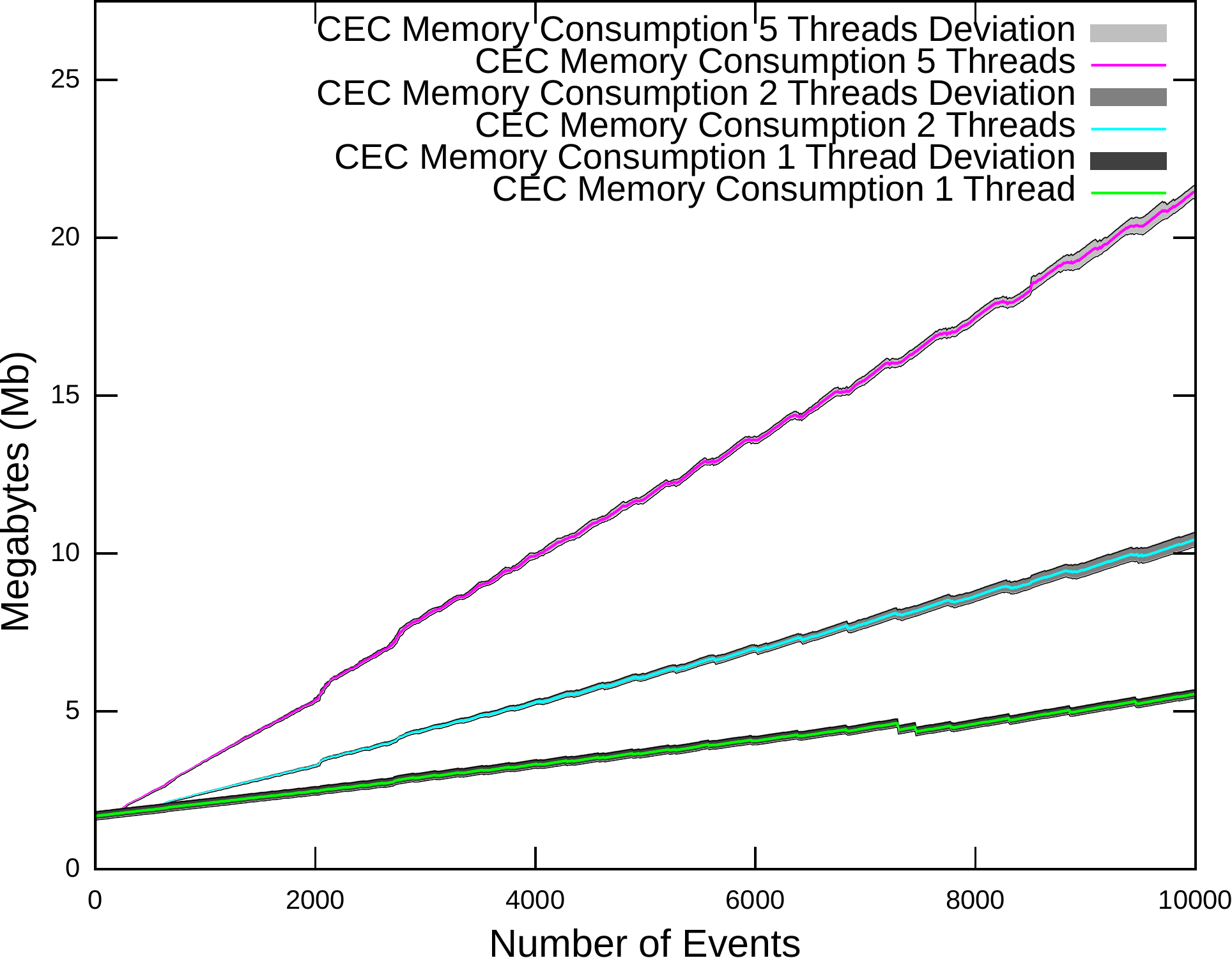}
							\caption{Effect of Multi-Threading on the Update Time and RAM Memory Consumption of CEC.}
							\label{multicec}
						\end{figure}
						
						On the top of Fig. \ref{queryupdate2} we show what happens when we perform a {\sf holds\_at/2} query where both the fluent and the value are variables and where we look for 
						all the available solutions. This is probably the heavies query for CEC as 2Prolog requires to perform a linear search on all the intervals to test whether these intervals intersect the time when the query is performed. For \ceckd~this query requires only to perform an 
						{\sf range\_query\_kdi/3} as shown in the previous section, which on side is proportional to the square root of the number of MVIs stored in the MVI tree, and on the other
						side it avoids accessing the MVIs by unification. 
						
						\begin{figure}[htp]
							\centering
							\includegraphics[width=6cm, height=4cm]{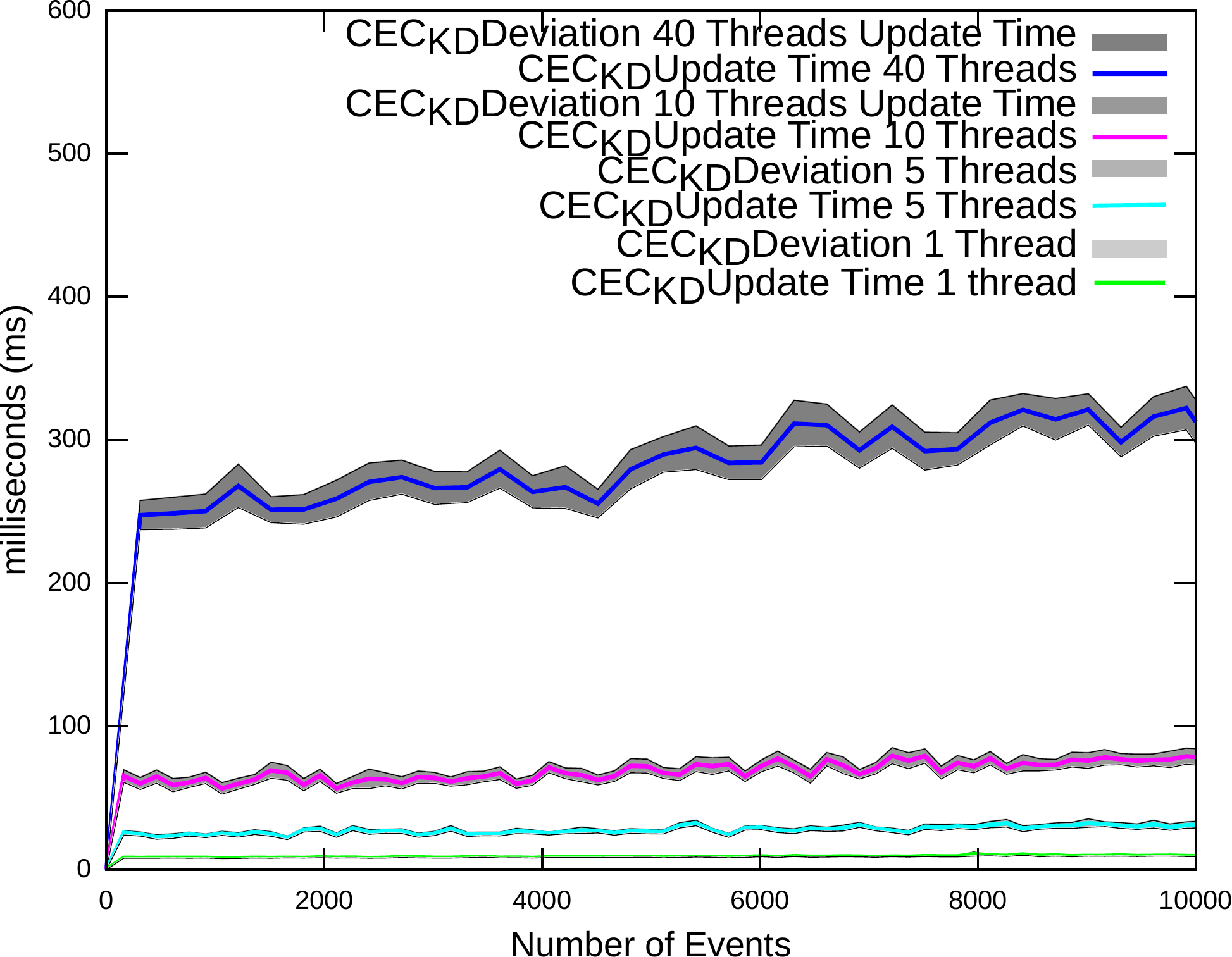}
							\includegraphics[width=6cm, height=4cm]{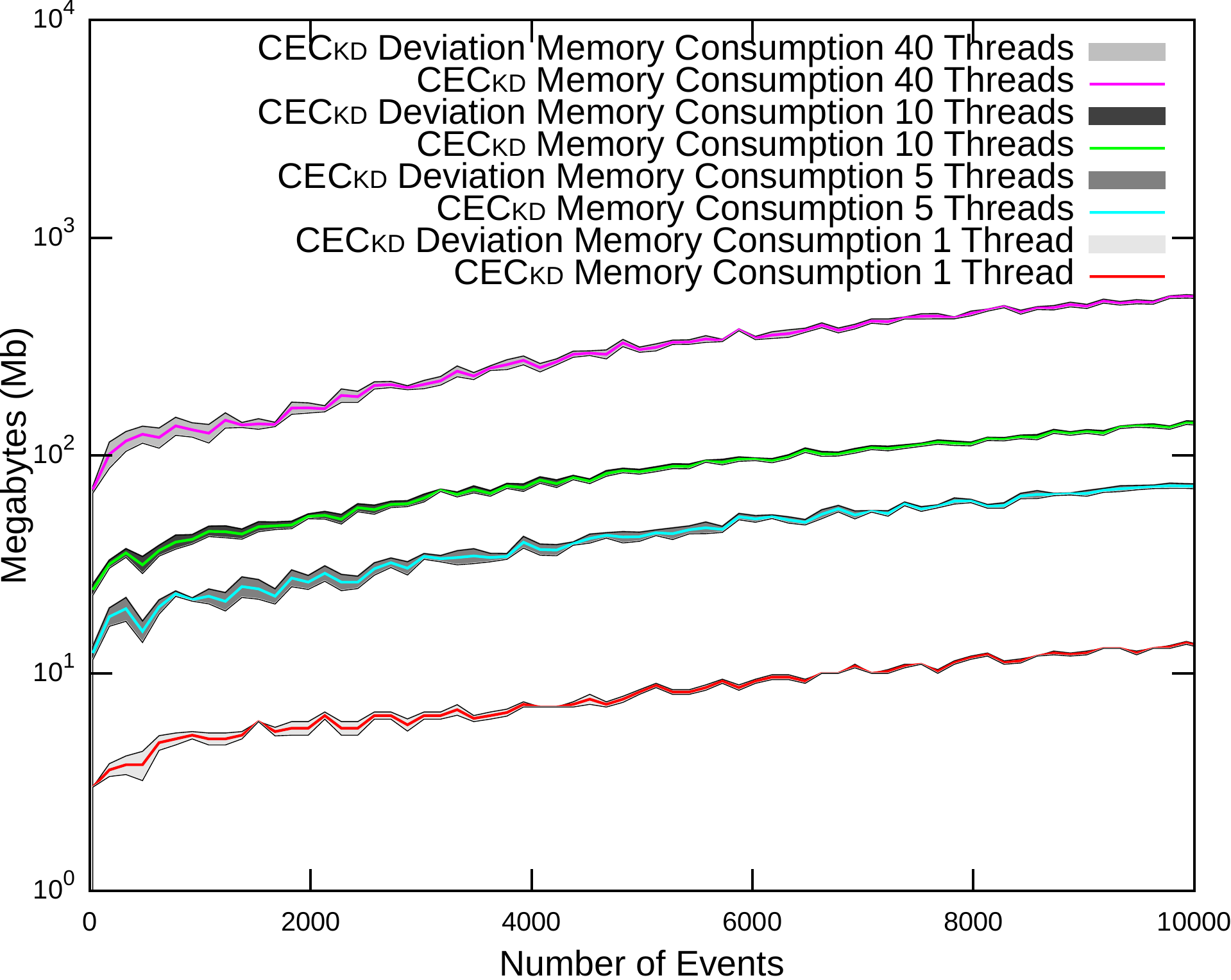}
							\caption{Effect of Multi-Threading on the Update Time and RAM Memory Consumption of \ceckd.}
							\label{multikcec}
						\end{figure}
						
						The part on the bottom of Fig. \ref{queryupdate2} illustrates the memory consumption of CEC and \ceckd~ in a monothread simulation. The result suggests that \ceckd~consumes twice as much RAM memory than CEC. This is explainable by the fact that we use two different trees for storing the events and the time periods and by the overhead introduced by the tree itself. 
						
						Fig. \ref{multicec} shows the effects of multi-threading on the computation of the update time in CEC. Despite running the tests on 2Prolog, which is optimised to run hundred of thousands of inferences, when adding multiple instances of the system loaded with CEC we have a situation where the instances compete for the CPU time, slowing themselves down, due to heavy use of backtracking and unification for each thread. 
						Since the amount of patients that a PHS may monitor may be very large, using CEC may require to use a big amount of independent machines, which would imply a very expensive solution for the medical doctors running the PHS. The fact that the RAM memory consumption of CEC is quite contained does not help in the multi-threaded tests on the top of Fig. \ref{multikcec}.
						
						As far as \ceckd\ is concerned, the use of Kd-trees allows our agents to avoid performing computationally demanding inferences using the 2Prolog engine and to query directly the Kd-trees containing the events and the MVIs of the fluents limiting the competition between the instances to obtain CPU time. In other words, the operation of accessing the nodes of the Kd-tree is less expensive than the unification procedure from the perspective of CPU usage, resulting in an efficient multi-threading behaviour in \ceckd.
						
						As shown on the bottom of Fig. \ref{multikcec} we can easily run 40 threads with 40 different \ceckd~instances and still having a quite acceptable update time. From the perspective of the memory consumption, the behaviour of \ceckd~is acceptable even when running 40 agents. 
						As shown on the right of Fig. \ref{multikcec}, 40 agents loaded with \ceckd~use around 500Mb of RAM, an amount available in most computers. Remembering that this is a crash test and that it is not the case that 40 medical doctors will use such an interface all at the same time, we conclude that \ceckd~is a good engine to support retrospective analysis of time series in the chronic non communicable diseases settings.

						\section{Related Work}
						\label{sec:relwork}
						
						This paper is both related with works in the fields of complex event processing (CEP) \cite{cugola2015introducing} and in the field of efficient computing in logic programming. Concerning CEP the combination of a graphical formalism with EC rules finds a close contribution in \cite{Cicekli:2006:FSE:2169225.2169259}, that uses the EC to model workflows. With respect to \cite{Cicekli:2006:FSE:2169225.2169259} the main contribution of this paper is that it models the logic formalism in blocks, using meta-programming techniques, where \cite{Cicekli:2006:FSE:2169225.2169259} rather focuses on execution problems concerning the workflow. Another prominent example is given by the ETALIS system \cite{AFRSSS2011}. ETALIS also defines how events evolve the state of a system, similarly to the EC. As a consequence of this similarity, it shares the same issues in terms of scalability. The main contribution of \ceckd~ with respect to ETALIS is that \ceckd has been developed to work on large streams of discrete and continuous events, whereas ETALIS focuses on discrete representations.
						Finally, another interesting contribution with respect to CEP with logic formalism is that provided by Teymourian et al in \cite{Teymourian2009}. Such a contribution presents the combination of a logic formalism with issues such as subsumption and classes of events. Currently, the semantic representation of \ceckd~ is flat, events do not have a semantic representation, in this sense producing an extension of \ceckd that also consider semantics could be an interesting development.
						
						There are a number of papers addressing efficient knowledge representation in the EC. We focus on those ones that keep the normal logic programming semantics as the original EC of ~\cite{ec}. For many practical applications we have found that the simple EC in ~\cite{DBLP:books/sp/wooldridgeV99/Shanahan99} with recursive predicate definitions in the rules is sufficient. Formalisms that are more expressive~\cite{DBLP:journals/logcom/Mueller04,DBLP:conf/ijcai/KimLP09} often constrain some uses of recursion and are therefore beyond the scope of this work.
						
						Before Chittaro and Montanari in~\cite{ChittaroM96,DBLP:conf/aime/ChittaroRD95}, Kowalski in ~\cite{DBLP:journals/jlp/Kowalski92} identified approaches to index events and validity intervals in EC influencing the research that took place subsequently. The Object Event Calculus (OEC)~\cite{DBLP:journals/tkde/KesimS96} can be seen as an extension of these ideas by relying on a simple version of the EC to model complex objects and their evolution in time. One of the attractive features of the OEC, is the ability to separate fluents representing single and multiple value attributes. Although the way these fluents were terminated by an event could be optimised, the overall knowledge representation required more axioms for the initiation of attributes that cater for the object-based data model and their underlying inheritance structure.
						
						Other attempts, such as the Reactive Event Calculus (REC), defined by Chesani et al. \cite{DBLP:journals/fuin/ChesaniMMT10}, builds on top of CEC, deriving the values of the maximum validity intervals by means of abductive reasoning on top of the SCIFF reasoner in order to achieve properties such as {\em irrevocability} and {\em soundness} of the EC. The axiomatisation of REC does not rely on {\sf assert} and {\sf retract} as in the other Prolog versions of the EC, but it relies on the constraint propagation mechanisms of SCIFF. This brings the advantage that the specification of REC is fully declarative, but, as reported in the developer notes of SCIFF\footnote{SCIFF developer Manual: http://lia.deis.unibo.it/sciff }, SCIFF indexes only on the functor of the produced events, which brings serious drawbacks on the computation time when dealing with large narratives.
						
						Urovi et al. in \cite{DBLP:conf/kr/UroviBSA10} present a scalability evaluation of the Ambient Event Calculus (AEC) \cite{golemdebs}, a distributed and indexed version of the OEC. Urovi et al. evaluate the AEC with many thousands of events considering local and distributed settings . We do not consider distributed settings, we rather focus on scaling up \ceckd\ for event recognition purposes, but, from the times reported in \cite{DBLP:conf/kr/UroviBSA10}, where for 5000 events an update time and a query time of 2 seconds for the local tests, we can say that \ceckd\ indexes better the events than AEC, which is based on the underlying indexing mechanisms of OEC.
						
						In \cite{Artikis:2010:LPA:1877937.1877941} Artikis et al. propose the LTAR-EC and they explain how the events are indexed by using a time window at compilation time.
						Our main differences with the Artikis et al. work is that we do not need to compile our theory with a fixed time window, we can specify, if needed, an arbitrarily wide time window at runtime. Similarly, we do not rely on the Prolog indexing capabilities as these may vary between different Prolog implementations, we rather use a Kd-tree indexing mechanism that allows us to represent our MVIs and events as multidimensional points of which we index several properties to speed up both the query and update time.

						A recent contribution to the problem of indexing EC events is also discussed in \cite{CI17}. Such a contribution defines as an activity recognition framework based on compiling the EC events happening within a certain time period, that comprises a month for the purpose of the application described in the paper. This certainly renders the EC more scalable, but it is dependent on the specific granularity of time selected.
						The most interesting aspect of \ceckd\ is that it handles time as a continuous attribute, supported by the capabilities of Kd-trees of handling spatio-temporal multi-dimensional points. As a consequence, \ceckd\ is time granularity invariant, although still dependent on the number of events happening, its indexing is not
						statically defined.

						\section{Conclusions and Future Work}
						\label{sec:conc}
						This paper presented the strategy that D1NAMO uses to analyse events from CGM devices and point of care devices. Specifically, in this paper we have presented an efficient version of the Event Calculus that caches events and their effects using an indexing scheme based on Kd-trees. We have studied the benefits of such integration by showing how to revisit previous work to produce a new temporal reasoning system that we called \ceckd. One of the main advantages of \ceckd\ is that it allows us to support scenarios that produce long narratives on which intelligent agents can reason about how to monitor patients. 
						We tested the formalism on events coming form D1NAMO dataset and compared the result with respect to the standard CEC, showing that our formalism outperforms the original CEC in terms of querying times, but implying a slight increase in the use of the RAM memory.
						Possible future work implies looking into the problem of creating an ontology of events, in order to allow description logic reasoning in terms of subsumption. Another possibility could be to change the indexing mechanism of \ceckd~ and try different structures than KD-trees. 
						
					\section{Acknowledgements}
					This work was partially funded by the FP7 Project COMMODITY12 Grant Agreement No. 287841 and by the Nano-tera.ch initiative through the D1NAMO project. We would like to acknowledge the precious comments of Professor Kostas Stathis concerning earlier versions of this paper.

\end{document}